\documentclass{article}

 \PassOptionsToPackage{numbers, compress}{natbib}
\usepackage[final]{nips_2016}

\usepackage[utf8]{inputenc} 
\usepackage[T1]{fontenc}    
\usepackage{hyperref}       
\usepackage{url}            
\usepackage{booktabs}       
\usepackage{amsfonts}       
\usepackage{nicefrac}       
\usepackage{microtype}      
\usepackage{amsmath,amssymb}     
\usepackage{mathtools}

\usepackage{xcolor}

\title{Transfer Learning Across Patient Variations with Hidden Parameter Markov Decision Processes}

\author{
   Taylor W.~Killian\\
   Harvard University \\
   \texttt{taylorkillian@g.harvard.edu} \\
   \And
   George Konidaris\\
   Brown University\\
   \texttt{gdk@cs.brown.edu}
   \And
   Finale Doshi-Velez\\
   Harvard University \\
   \texttt{finale@seas.harvard.edu}
}

\begin{document}

\maketitle

\begin{abstract}
Due to physiological variation, patients diagnosed with the same
condition may exhibit divergent, but related, responses to the same
treatments.  Hidden Parameter Markov Decision Processes (HiP-MDPs)
tackle this transfer-learning problem by embedding these tasks into a
low-dimensional space. However, the original formulation of HiP-MDP
had a critical flaw: the embedding uncertainty was modelled
independently of the agent's state uncertainty, requiring an unnatural
training procedure in which all tasks visited every part of the state
space---possible for robots that can be moved to a particular
location, impossible for human patients.  We update the HiP-MDP
framework and extend it to more robustly develop personalized medicine
strategies for HIV treatment.
\end{abstract}

\section{Introduction}\label{sec:intro}


Due to physiological variation, patients diagnosed with the same
condition may exhibit divergent, but related, responses to the same
treatments.
To develop optimal treatment or control policies for a patient, it is
undesirable and ineffectual to start afresh each time a new individual
is cared for. However, each patient may still require a tailored
treatment plan as ``one-size-fits-all" treatments can introduce more
risk in aggressive diagnoses. Ideally, an agent tasked with developing
an optimal health management policy would be able to leverage the
similarities across separate, but related, instances while also
customizing treatment for the individual. This paradigm of learning
introduces a compelling regime for transfer learning.

The Hidden Parameter Markov Decision Process
({\small HiP-MDP})~\cite{doshivelez2013HiP_MDP} formalizes the transfer
learning task in the following way: first it assumes that any task
instance can be fully parameterized by a bounded number of latent
parameters {\small $w$}.  That is, we posit that the dynamics dictating a
patient's physiological response can be expressed as {\small $T(s' | s, a,
w_b)$} for patient {\small $b$}. Second, we assume that the system dynamics will
not change during a task and an agent would be capable of determining
when a change occurs (e.g. a new
patient). \citet{doshivelez2013HiP_MDP} show that the {\small HiP-MDP} can
identify the dynamics of a new task instance and flexibly adapt to the
variations present therein.  However, the original {\small HiP-MDP} formulation
had a critical flaw: the embedding uncertainty of the latent parameter
space was modelled independently from the agent's state
uncertainty. This assumption required the agent to have the ability to visit every part of the state
space before identifying the variations present in the dynamics of the
current instance. While this may be feasible in robotic systems, it is not generally available to domains in healthcare.

We present an alternative {\small HiP-MDP} formulation that alleviates this
issue via a Gaussian Process latent variable model ({\small GPLVM}). This
approach creates a unified Gaussian Process ({\small GP}) model for both
inferring the transition dynamics within a task instance but also in
the transfer between task instances~\cite{cao2010adaptive}. Steps are
taken to avoid negative transfer by selecting the most representative
examples of the prior instances with regards to the latent parameter
setting. This change in the model allows for better uncertainty
quantification and thus more robust and direct transfer. We ground our
approach with recent advances in the use of {\small GPs} to approximate
dynamical systems and in transfer learning as well as discuss relevant
Reinforcement Learning (RL) applications to healthcare
(Sec.~\ref{sec:prior_work}). In Sec.~\ref{sec:model} we formalize the
adjustments to the {\small HiP-MDP} framework and in Sec.~\ref{sec:experiments}
we present the performance of the adjusted {\small HiP-MDP} on developing personalized
treatment strategies within HIV simulators.

\section{Related Work}\label{sec:prior_work}

The use of RL (and machine learning, in general) in the development of
optimal control policies and decision making strategies in
healthcare~\cite{shortreed2011informing} is gaining significant
momentum as methodologies have begun to adequately account for
uncertainty and variations in the problem space. There have been
notable efforts made in the administration of
anesthesia~\cite{moore2014reinforcement}, in personalizing
cancer~\cite{tenenbaum2010personalizing} and HIV
therapies~\cite{ernst2006clinical} and in understanding the causality
of macro events in diabetes
managment~\cite{merck2015causal}. \citet{marivate2014quantifying} formalized a routine to
accommodate multiple sources of uncertainty in batch RL methods to
better evaluate the effectiveness of treatments across a
subpopulations of patients. We similarly attempt to address and
identify the variations across subpopulations in the development
treatment policies. We instead, attempt to account for these
variations while developing effective treatment policies in an
approximate online fashion.

{\small GPs} have increasingly been used to facilitate methods of
RL~\cite{rasmussen2003gpinrl, rasmussen2006gp4ml}. Recent advances in
modeling dynamical systems with {\small GPs} have led to more efficient and
robust formulations~\cite{deisenroth2011pilco,
  deisenroth2015gaussian}, most particularly in the approximation and
simulation of dynamical systems. The {\small HiP-MDP} approximates the
underlying dynamical system of the task through the training of a
Gaussian Process dynamical model~\cite{deisenroth2012expectation,
  wang2005gpdm} where only a small portion of the true system dynamics
may be observed as is common in partially observable Markov Decision
Processes (POMDP)~\cite{kaelbling1998planning}. In order to facilitate
the transfer between task instances we embed a latent, low-dimensional
parametrization of the system dynamics with the states. By virtue of the
{\small GP}~\cite{lawrence2004gplvm, urtasun2007discriminative}, this latent
embedding allows the {\small HiP-MDP} to infer across similar task instances
and provide a better prediction of the currently observed system.

The use of {\small GPs} to facilitate the transfer of previously learned information to new instances of the same or a similar task has a rich history~\cite{bonilla2008multitask, kaelbling1998planning, rasmussen2003gpinrl}. More recently, there have been advances in organizing how the {\small GP} is used to transfer, being constrained to only select previous task instances where positive transfer occurs~\cite{cao2010adaptive, leen2011focused}. This adaptive approach to transfer learning helps to avoid previous instances that would otherwise negatively affect effective learning in the current instance. By selecting the most relevant instances of a current task for transfer, learning in the current instance becomes more efficient. 

\section{Model}
\label{sec:model}

The {\small HiP-MDP} is described by a tuple: {\small $\left\{S , A , \Theta , T , R , \gamma , P_{\Theta}\right\}$}, where {\small $S$} and {\small $A$} are the sets of states {\small $s$} and actions {\small $a$} (eg. patient health state and prescribed treatment, respectively), and {\small $R(s,a)$} is the reward function mapping the utility of taking action $a$ from state $s$. The transition dynamics {\small $T\left(s' | s, a, \theta_b\right)$} for each task instance {\small $b$} depends on the value of the hidden parameters {\small $\theta_b\in\Theta$} (eg. patient physiology). Where the set of all possible parameters {\small $\theta_b$} is denoted by {\small $\Theta$} and where {\small $P_\Theta$} is the prior over these parameters. Finally, {\small $\gamma\in(0,1]$} is the factor by which {\small $R$} is discounted to express how influential immediate rewards are when learning a control policy. Thereby, the {\small HiP-MDP} describes a \textit{class} of tasks; where particular instances of that class are obtained by independently sampling a parameter vector {\small $\theta_b\in\Theta$} at the initiation of a new task instance {\small $b$}. We assume that {\small $\theta_b$} is invariant over the duration of the instance, signaling distinct learning frontiers between instances when a newly drawn {\small $\theta_{b'}$} accompanies observed additions to {\small $S$} and {\small $A$}. 

The {\small HiP-MDP} presented in~\cite{doshivelez2013HiP_MDP} provided a transition model of the form: 
\begin{small}
\begin{align*}
(s'_d - s_d) &\sim \sum_k^K z_{kad}w_{kb}f_{kad}(s)+\epsilon\\
\epsilon &\sim \mathcal{N}(0,\sigma^2_{nad})
\end{align*}
\end{small}
which sought to learn weights {\small $w_{kb}$} based on the {\small $k^{th}$} latent factor corresponding to task instance {\small $b$}, filter parameters {\small $z_{kad}\in\{0,1\}$} denoting whether the {\small $k^{th}$} latent parameter is relevant in predicting dimension {\small $d$} when taking action {\small $a$} as well as task specific basis functions {\small $f_{kad}$} drawn from a {\small GP}. While this formulation is expressive, it presents a problematic flaw when trained. Due to the independence of the weights {\small $w_{kb}$} from the basis functions {\small $f_{kad}$}, training the {\small HiP-MDP} requires canvassing the state space $S$ in order to infer the filter parameters {\small $z_{kad}$} and learn the instance specific weights {\small $w_{kb}$} for each latent parameter. 

We bypass this flaw by applying a {\small GPLVM}~\cite{lawrence2004gplvm} to jointly represent the dynamics and the latent weights $w_b$ corresponding to a specific task instance $b$. This leads to providing as input to the {\small GP}, with hyperparameters $\psi$, the augmented state $\tilde{s} =: [s^\intercal,\ a, \ w_b]^\intercal$. The approximated transition model then takes the form of:
\begin{small}
\begin{align*}
s'_d &\sim f_d(\tilde{s}) + \epsilon\\
f_d &\sim GP(\psi)\\
w_b &\sim \mathcal{N}\left(\mu_b, \Sigma_b\right) \\
\epsilon &\sim \mathcal{N}\left(0,\sigma_{bd}\right)
\end{align*}
\end{small}
This approach enables the {\small HiP-MDP} to flexibly infer the dynamics of a new instance by virtue of the statistical similarities found in the learned covariance function between observed states of the new instance and those from prior instances. Another feature of formulating the {\small HiP-MDP} after this fashion is that we are able to leverage the marginal log likelihood of the {\small GP} to optimize the weight distribution and thereby quantify the uncertainty~\cite{quinonero2004phd, quinonero2003gp} of the latent embedding of {\small $w_b$} for {\small $\theta_b$}. These two features of reformulating the {\small HiP-MDP} as a {\small GPLVM} allows for more robust and efficient transfer. 

\begin{figure}[h!]
\begin{centering}
  \includegraphics[width=0.4\linewidth]{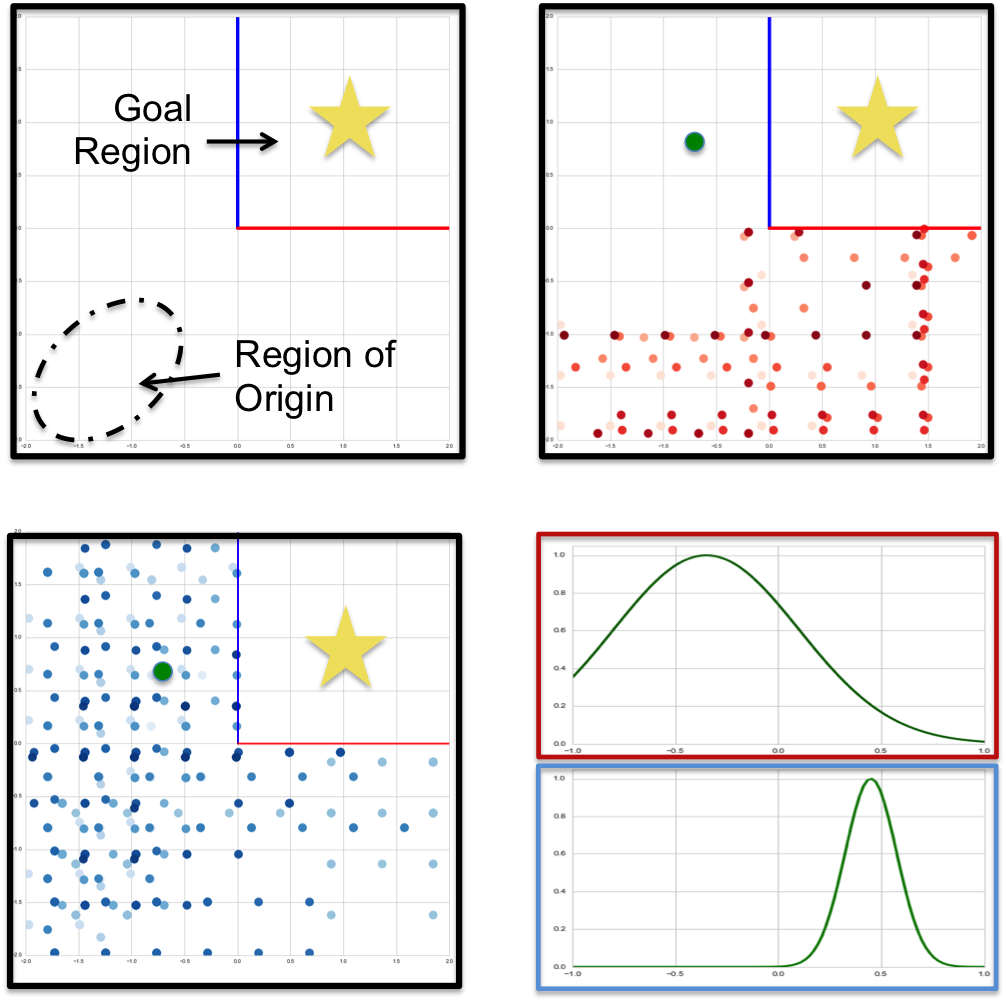}
  \caption{Toy Problem: (a) Schematic outlining the domain, (b) learned policy for ``red" parametrization, (c) learned policy for ``blue" parametrization, (d) uncertainty measure for input point according to separate latent classes.}
  \label{fig:demo}
\end{centering}
\end{figure}

\paragraph{Demonstration} We demonstrate a toy example {\small(see Fig. \ref{fig:demo})} of a domain where an agent is able to learn separate policies according to a hidden latent parameter. Instances inhabiting a ``blue" latent parametrization can only pass through to the goal region over the blue boundary while those with a ``red" parametrization can only cross the red boundary. After a few training instances, the {\small HiP-MDP} is able to separate the two latent classes and develops individualized policies for each. Due to the flexibility enabled by embedding the latent parametrization into the system's state, the {\small GPLVM} identifies which class the current instance belongs to within the first couple of training episodes. In total, this example took approximately 30 minutes to develop optimal policies for 20 task instances. We place an unclassified survey point in the top left quadrant to gather information about the policy uncertainty given the two latent classes.

\section{Inference}
\label{sec:inference}

\paragraph{Parameter Learning and Updates} We deploy the {\small HiP-MDP} when the agent is provided a large amount of batch observational data from several task instances (e.g. patients) and tasked with quickly performing well on new instances. With this observational data the {\small GP} transition functions $f_d$ are learned and the individual weighting distributions for $w_b$ are optimized. However, the training of the $f_d$ requires computing inverses of matrices of size $N = \sum_b n_b$ where $n_b$ is the number of data points collected from instance $b$. To streamline the approximation of $T$ we choose a set of support points $s^*$ from $S_b$ that sparsely approximate the full {\small GP}. Optimization procedures exist to select these points accurately~\cite{rasmussen2006gp4ml, snelson2005sparse}, we however heuristically select these points to minimize the maximum reconstruction error within each batch using simulated annealing.

\paragraph{Control Policy} 
A control policy is learned for each task instance $b$ following the procedure outlined in~\cite{deisenroth2011pilco} where a set of tuples $\left(s, \ a, \ s', \ r\right)$ are observed and the policy is periodically updated (as is the latent embedding $w_b$) in an online fashion, leveraging the approximate dynamics of $T$ via the $f^*_d$ to create a synthetic batch of data from the current instance. This generated batch of data from $b$ is then used to improve the current policy via the Double Deep Q Network variant of fitted-Q using prioritized experience replay~\cite{mnih2013playing, schaul2015prioritized, van2015deep}. Multiple episodes are run from each instance $b$ to optimize the policy for completing the task under the hidden parameter setting $\theta_b$. After doing so, the hyperparameters of the {\small GP} defining the $f_d$ are updated before learning for another randomly manifest task instance.

\section{Experimental Results}
\label{sec:experiments}

\paragraph{Baselines}

We benchmark the {\small HiP-MDP} framework in the HIV domain by observing how an agent would perform without transferring information from prior patients to aid in the efficient development of the treatment policy for a current patient. We do this by representing two ends of the precision medicine spectrum; a ``one-size-fits-all" approach that learns a single treatment policy for all patients by using all previous patient data together and a "personally tailored" treatment plan where a single patient's data is all that is used to train the policy. We represent these baselines in environments where a model is present (with the simulators) or absent (utilizing the {\small GP} approximation).

\paragraph{HIV Treatment}

\begin{figure}[h!]
\begin{centering}
  \includegraphics[width=0.565\linewidth]{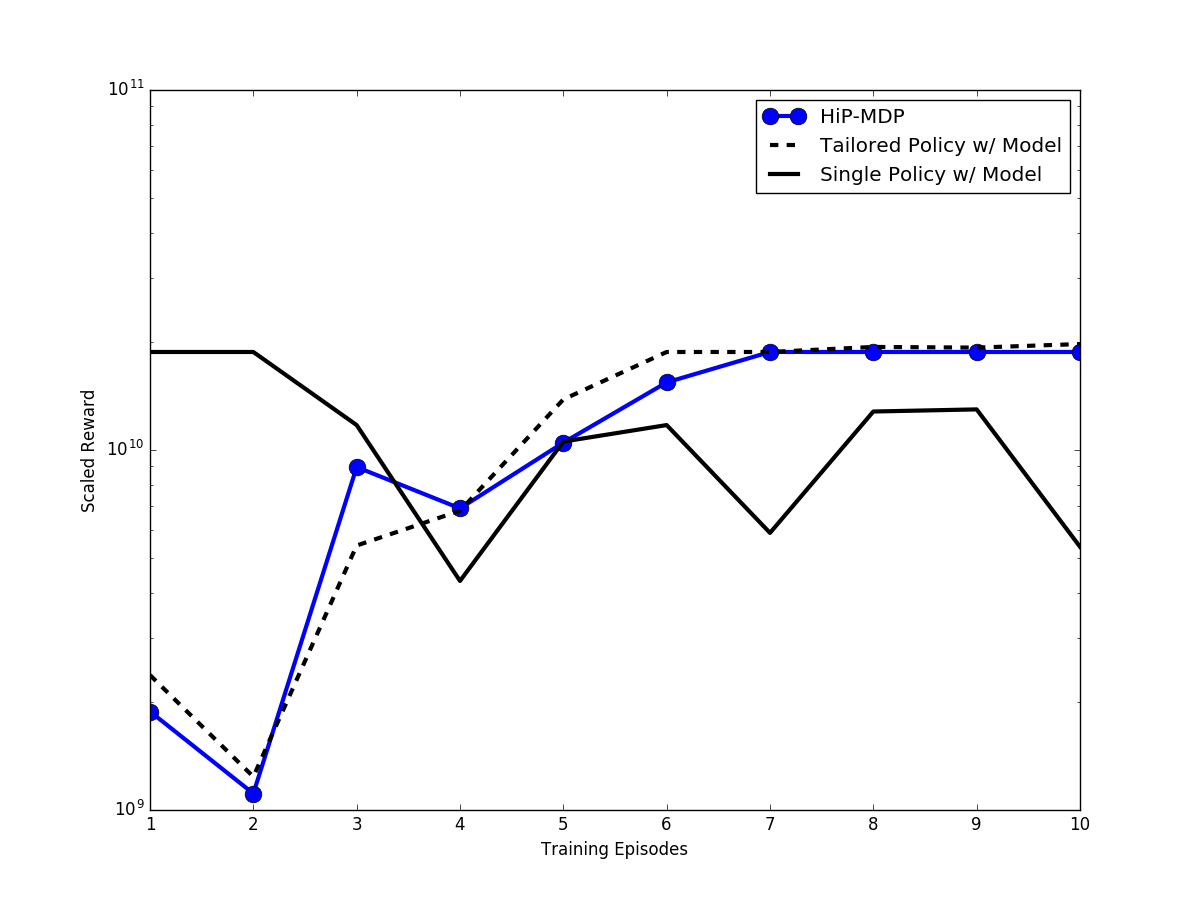}
  \caption{Representative results of applying the {\small GPLVM} aided {\small HiP-MDP} model to the HIV treatment simulator as provided from~\cite{ernst2006clinical}. The {\small HiP-MDP} learned treatment policy (blue) matches or improves on the naive baseline policy development strategies.}
  \label{fig:results}
\end{centering}
\end{figure}

Ernst, et.al.~\cite{ernst2006clinical} leverage the mathematical representation of how a patient responds to HIV treatments~\cite{adams2004dynamic} in developing an RL approach to find effective treatment policies using methods introduced in~\cite{ernst2005tree}. The learned treatment policies cycle on and off two different types of anti-retroviral medication in a sequence that maximizes long-term health. By perturbing the underlying system parameters one can simulate varied patient physiologies. We leverage these variations via the {\small GPLVM} augmentation to the {\small HiP-MDP} to efficiently learn treatment policies that match the naive ``personally tailored" baseline but with reliance on much less data. The {\small HiP-MDP} also outperforms the ``one-size-fits-all" baseline, as expected. {\small(see Fig. \ref{fig:results} for representative results)}. We see that the {\small GPLVM} driven {\small HiP-MDP} is capable of immediately taking advantage of the prior information from previously learned data even in the face of unique physiological characteristics. The robust and efficient manner in which the {\small HiP-MDP} achieves such results in the HIV domain is promising and, in turn, motivates further inquiry into a more generalized learning agent for the development of other individualized medical treatment plans.

%
%
%
%
%
%


\bibliographystyle{plainnat}
\bibliography{bibliography}

\end{document}